\documentclass[conference]{IEEEtran}
\usepackage{cite}
\usepackage{amsmath,amssymb,amsfonts}
\usepackage{algorithmic}
\usepackage{graphicx}
\usepackage{textcomp}
\usepackage{xcolor}
\usepackage{url}
\usepackage{float}
\usepackage{cuted}
\usepackage{graphicx, caption}

\begin{document}

\title{Deep Learning Inference Frameworks Benchmark}

\author{\IEEEauthorblockN{Pierrick Pochelu} \\
pierrick.pochelu@gmail.com
}

\maketitle

\begin{abstract}
Deep learning (DL) has been widely adopted those last years but they are computing-intensive method. Therefore, scientists proposed diverse optimization to accelerate their predictions for end-user applications.  However, no single inference framework currently dominates in terms of performance. This paper takes a holistic approach to conduct an empirical comparison and analysis of four representative DL inference frameworks. First, given a selection of CPU-GPU configurations, we show that for a specific DL framework, different configurations of its settings may have a significant impact on the prediction speed, memory, and computing power. Second, to the best of our knowledge, this study is the first to identify the opportunities for accelerating the ensemble of co-localized models in the same GPU. This measurement study provides an in-depth empirical comparison and analysis of four representative DL frameworks and offers practical guidance for service providers to deploy and deliver DL predictions.
\end{abstract}


\section{Introduction}
\label{sec:infersys}

% What is an inference framework?
An inference deep learning framework consists in predicting with an already-trained neural network. After training, the inference frameworks TensorRT \cite{tensorrt}, ONNX-runtime  \cite{onnx:}, OpenVINO \cite{openvino}, Tensorflow XLA \cite{xla}, LLVM MLIR \cite{mlir} apply diverse optimizations to accelerate its computing speed.

% Why it is important?
The last decade shows that bigger deep learning models are generally more accurate. However, they are also slower and memory cumbersome. Accelerating their predictions is, therefore, crucial for speed-critical applications, and applications where prediction quality is the priority with reasonable speed. While the training of DNNs is a time-bounded activity, the inference is generally deployed and runs for a long time. This is why optimizing the prediction time of deep neural networks is sometimes considered more strategic than the training \cite{aws:2019}.

% But du papier
This paper analyzes the prediction performance of some frameworks: Tensorflow, ONNX, OpenVINO, and TensorRT benchmarked on diverse computer vision neural networks. For each framework, we gather more than 80 values including throughputs (predictions per second), load time, and memory consumption, power consumption on both GPU and CPU. We provide a comprehensive review of used optimization technics and evaluate them to guide the inference framework choice. The figures are presented in appendices and the link to the code is given at the conclusion.

% GAP in the literature
Some benchmarks have been proposed to measure deep neural network speeds. Some proposed microbenchmarks such as AI-Matrix\cite{aimatrix} aim to measure the speed of operators such as matrix multiplication and convolution, but they are far from evaluating the overall complexity of modern neural networks. Dawn Bench \cite{dawn:2019} shows that neural networks underutilize cores due to the bottleneck of the memory transfers. Benchmarks such as MLPerf Inf \cite{mlperfinf}, and ML Bench \cite{mlbench} are comprehensive studies of different inference applications, inference frameworks, and hardware. However, they generally do not provide an in-depth analysis of the inference framework settings.

% Ma contrib
First, given a selection of two computing nodes, the deep learning model, and the inference framework settings are analyzed. We measure that for a specific neural network, different inference frameworks may have a significant impact on prediction speed, memory, and computing power. Second, to the best of our knowledge, this study is the first to identify the opportunities for accelerating ensembles of co-localized neural networks in the same GPU which is a more and more performed deep learning procedure \cite{fleet} \cite{mldb:2021}. This measurement study provides an in-depth empirical comparison and analysis of four representative DL frameworks and offers practical guidance for deploying and delivering DL predictions to the end-user application.

% Presenter le papier
Section~\ref{sec:opt} presents the different representations and optimization methods of the inference frameworks. Next, section~\ref{sec:api} describes a common API to perform many comparisons and analyses easier. Section~\ref{sec:settings} explicit the settings and enabled optimizations of each framework. Section~\ref{sec:res}, the results of the experiments break down into four categories: computing time, memory consumption, power consumption, and loading time. For the first, time benchmarks of an ensemble of neural networks predicting together are included.
The conclusion in section~\ref{sec:conclusion} summarize the results, provides insight for the future, and the link for the code.

\section{Post-training representation and optimization}
\label{sec:opt}
Computational graphs provide a global view of operators (nodes) and tensor exchanges (edges) by avoiding specifying
how each operator must be implemented. Operators are generally implemented in a low-level code representation.

The post-optimization is an intermediate step between training and inference to optimize the model representation. It generally exploits this high-level representation to change the computational graph, and the code representation to compile and optimize the low-level code targetting the specific hardware.

These high-level and low-level optimizations are generally orthogonal to global optimization, and they can be used together to obtain further performances.

\textbf{High-level optimization.} \textit{Fusing} consists of merging multiple operations in one single kernel launch. For example, it is possible to merge a sequence of convolutions \cite{groupedconv:2020} and a sequence of dense operations such as ``Y=AX+B''. 

Fusing removes the need for slow storage of some intermediate tensors and improves cache utilization. It also removes the synchronization barriers between operations maximizing the utilization of the cores. We may notice that those optimization types may be also beneficial to the training phase such as XLA \cite{xla}.

And more, some fusing operations consist of mathematical simplification. For example, the adjacent convolution layer and batch normalization layer may be merged into one single convolution layer with changed weights \cite{batchconv:2020}.

Other high-level optimizations include \textit{constant-folding} and \textit{static memory planning} which pre-computes graph parts operations and intermediate tensor buffers.

\textbf{Low-level optimization.} Converting the computing graph into a low-level code optimized for targeted hardware. Those optimization benefits from decades of knowledge of compilation technics. It includes sub-expression elimination, vectorization, loop ordering/tiling/unrolling, threading pattern, and memory caching/re-using ... Authors propose TVM \cite{tvm:2018} and LLVM MLIR \cite{mlir} low-level compilers and optimizers enriched with the tensor type.

% Produced code contains the tensor type, hardware intrinsic

%It allows the application of compilation optimization technics. We may enumerate 

%The combinatorial choices of memory access, threading pattern, and novel hardware primitives
%creates a huge configuration space for generated code.

%Tiling with automatic tuning of the tile size according to the hardware

%nested parallelism, tensorization Special memory-scope enables memory reuse in GPUs and explicit management of on-chip memory in accelerators

%Automatic tuning of the map \cite{halide:2016} between tensor expression to low-level code based on compute/schedule.
% https://www.usenix.org/system/files/osdi18-chen.pdf

%TODO citer LLVM MLIR, citer TVM

%\textbf{Pruning}

%\textbf{Lower precision arithemetics} % They aim for a reduction in memory and on the disk, ideal for the big deep neural network. Constructors have recently proposed different formats to balance between

% EXpliquer le fusing, pruning, code optimization (mimalloc, ) , lower precision.

% Batch size in deep learning
A critical part of optimizing the performance of a DNN model is its batch size. It controls the internal cores utilization, the memory consumption, and the data exchange between the CPU containing input data and the device supporting the DNN (if different). The best batch size value is unpredictable, therefore it is a common practice to scan a range of potential values.

%Some optimize them with optimize.  the effect of the batch size and adapt the batch automatically Short batch size accelerates. Some inference system try . This is why some inference servers tune 

%Some tools, allow scanning different batch sizes automatically and return the best performing one. Some inference systems optimize and tune automatically inference framework settings using greedy approach \cite{pi}. Inference servers are generally adapted to be 

%\subsection{Inference frameworks}

%The ``load'' function loads a trained DNN from the disk to a targeted device and the prediction function $f(x) \rightarrow y$ with $x$ the data samples and $y$ the associated predictions. 

%The most sophisticated inference frameworks \cite{tensorrt}, \cite{openvino}, \cite{onnx:}, \cite{tflite:} perform post-training optimization such operations optimization and device-specific optimization with low or no impact on the accuracy. We use here the Tensorflow deployment (``pb files'') as the underlying format and focus our work on the allocation challenge. 

\section{Inference frameworks unified API}
\label{sec:api}

Different inference frameworks have different API, this is why a common Python API is proposed here. The overhead of calling this function is negligible. It contains two functions ``$load(path,params)$'' and ``$predict(x) \rightarrow y$''.

``$load(path, params)$'' function loads a previously trained model from the disk at the location $path$ and applies post-training with the dictionary parameters $params$. The optimized DAG is stored on the targeted GPU given by $params[`gpuid']$ and `-1' for the CPU.

Since post-optimization may take minutes, inference frameworks generally propose caching the optimized DAG on the disk. One application of this cached-optimized DAG is the elasticity of the prediction as a micro-service. % of cloud The optimized DAG is faster to load and may allow elasticity.

The prediction function is given with $f(x) \rightarrow y$ with $x$ the data samples and $y$ the associated predictions. $x$ is split into batches of size $params[`batch\_size']$. The batching is done asynchronously in $f$ to overlap the computing of the prediction with the DAG and the movements of $x$ and $y$.

%\subsection{Inference servers}

% Machine Learning inference server
%The last 10 years we show the emergence of machine learning inference servers to serve trained models as a service \cite{laser:2014} \cite{clipper:2017} \cite{fang:2017} \cite{dbms} aiming for performance measured with the throughput (predictions-per-second under a heavy workload) or latency (time to answer to a request). 

% Les servers
%The last few years, we show the emergence of software \cite{clipper:2017} \cite{tritonserv}, \cite{ray}, \cite{tfserv} to serve inference framework predictions as a service. Most of them serve wrap prediction in a REST service or sometimes in a database management service \cite{dbms}. They implement often the same features. The model selection allows the client application to choose the model which will answer among multiple applications or the same application but the desired trade-off between accuracy and speed. To improve performance under redundant requests, caching allows for avoiding recomputing similar requests. When the amount of requests is low and irregular, adaptative batching allows triggering prediction before the buffered batch is full to improve the latency. We will not enter the details of the engineering of the server to stay focused on the prediction of DNNs.

 %Figures show that the directed acyclical graph (DAG) is generally bottlenecking the performance or slow data movements.

\section{Experimental settings}
\label{sec:settings}
%Like previously seen in section~\ref{sec:operations}. However, they are multiple expected benefits to using a lower precision format: faster arithmetic operations, faster memory exchange, lower memory consumption, and lower power consumption. This is why lower precision has been introduced under different arithmetic formats and different procedures to increase the training and the inference phase.
%Mixed precision store the parameters in FP16 but it computes FP32 gradients 
%to keep precision when deployed.
%e recommend maintaining a single-precision copy of weights that accumulates
%the gradients after each optimizer step (this copy is rounded to half-precision for
%the forward- and back-propagation). Secondly, we propose loss-scaling to keep gradient values with small magnitudes. Thirdly, we use half-precision arithmetic that accumulates into single-precision outputs, which are converted to half-precision before storing in memory

\textbf{Neural networks.} ML scientists have developed a variety of convolutional blocks (e.g., VGG, Residual connect). They propose different neural networks (e.g., Resnet50, Resnet101, ...) with sometimes the help of automatic tuning technics to optimize and tune the trade-off between accuracy and computing efficiency. We present four of them in table~\ref{tab:topology}.

\begin{table}[h]
\small
\setlength\tabcolsep{1.5pt}
\caption{Four deep neural network architecture. We choose them diverse in terms of depth, width (rate between \#params and \#layers), and density.}
\label{tab:topology}
\begin{tabular}{llllll}
               & $\frac{\#param.}{\#layers}$ & \#layers & \#jumps & \#param. & Jump type           \\
VGG19          & 7.26M                                                             & 19                                                       & 0                                                         & 138M   & N/A                 \\
ResNet50       & 0.52M                                                             & 50                                                       & 16                                                        & 26M    & Additions           \\
DensetNet201   & 0.1M                                                              & 201                                                      & 98                                                        & 20M    & Concatenations      \\
EfficientNetB0 & 0.06M                                                             & 89                                                       & 25                                                        & 5.3M   & Mult. and Add.
\end{tabular}
\end{table}

% Ensembles
%While most of the benchmark evaluates a few operations or some individual neural networks, the current trend in the ML community seems to be interested in deploying multiple DNNs at the same time for diverse reasons: Independent DNN prediction services,  ensemble of neural network to improve predictions over one single model, Multi-agent reinforcement learning, cascading neural networks, ...

GPU technologies have evolved in computing devices containing thousands of cores and gigabytes of memory, and modern deep neural networks may not guarantee the use of all resources in one single GPU.  To match the ML community computing needs, an ideal inference framework should not only efficiently leverage its resources to one DNN but should be performant to run multiple DNNs at the same time. That is why propose the ensembles in our thesis. All values in parentheses are our evaluated test top1-accuracy on ImageNet: VGG19 (69.80\%), ResNet50 (74.26\%), DenseNet201 (75.11\%) and ensemble them by interpolating their predictions performs 76.92\%. We choose not to add efficient-netB0 in the ensemble due to operations errors with some frameworks.

Our benchmarks are performed on two machines. We do not distribute inference over multiple GPUs for keeping the analysis focused on the inference framework. Even if we make expect performance gain using the map-reduce paradigm. %Evaluations are made on only one GPU or the CPU at a time.

\textbf{Machine A} is an HGX1 equipped with Tesla V100 SXM2 GPUs containing 5120 Cuda cores running at 1312Mhz-1530Mhz frequency and 16G of GPU memory. Its CPU is a dual-socket Intel(R) Xeon(R) CPU E5-2698 v4 with a total of 80 cores @ 1.2Ghz-3.6Ghz and 512GB of RAM. The total GPU board power is 300 watts.

\textbf{Machine B} is an in-house machine equipped with NVIDIA Amper A100 PCI-E GPUs, each one containing 6912 Cuda cores running at 765Mhz-1410Mhz and 40G of GPU memory.  Its CPU is a single-socket AMD EPYC 7F52 with 16 cores running at 2.5Ghz-3.5Ghz and 256GB of RAM. The total GPU board power is 250 watts. 

\textbf{Software stack.} All software stack versions are the same on both machines: machine A and machine B. The assessed inference framework versions are Tensorflow 2.6, TensorRT 8.0, Onnx-runtime 1.10, and OpenVINO 2021.  The needed neural network converters to benchmark those frameworks are tf2onnx 1.9.3, and LLVM 14.0.

Machine A is running on Ubuntu and we use Python 3.9, Machine B is running on CentOS and uses Python 3.8.  Specialized OS for real-time may improve latency determinism and latency speed but could lower throughput performance too. We evaluated MLIR (onnx-mlir 0.2 framework \cite{mlir}) on Machine B but it support only ResNet50.

In all our benchmarks Tensorflow is accelerated with XLA \cite{xla} (Accelerated Linear Algebra).

\textbf{Graph optimization settings:}
\begin{itemize}
    \item \textbf{Tensorflow XLA}: The computing graph is frozen (i.e., all weights are put in ``read-only memory''). The optimizer "optimize\_for\_inference\_lib" is enabled but it does not impact the performance. XLA is enabled on GPU, disabling it reduces speed by 15\%. However, enabling XLA multiplies the initialization time by factor 6. And more, we do not observe performance gain to enable it on the CPU so we let it be disabled on the CPU.
    \item \textbf{ONNX-RT} (ONNX-runtime) \cite{onnx:}: Caching is enabled, disabling it reduces speed by ~3\%. The maximum graph level optimization is enabled, and using the default optimization settings reduces speed by 8\%.
    \item \textbf{OpenVINO} \cite{openvino}: The ``NCHW'' tensor format (images, channels, height, and width) is enabled meaning all images must be converted in this format. The convolution fusing is enabled, disabling it reduces speed by 9\%. The concatenation optimization is disabled because it would hurt performance by 1\% on DenseNet201. 
    \item \textbf{MLIR} \cite{mlir} (Multi-Level Intermediate Representation from the LLVM project) is still under development. The graphs are compiled with the ``-O3'' option.
\end{itemize}

\section{Experimental results}
\label{sec:res}

Not all metrics are equally important for all ML applications, and weighting some more heavily than others is highly application-dependent. This is why we bench multiple metrics: prediction speed, memory consumption, power consumption, and loading time of the model.

\subsection{Prediction speed}

We identify three main scenarios and associated speed measures:
\begin{itemize}
    \item \textbf{Batch applications.} First, when the neural network predicts a large workload of data samples, the batch size must be optimized to provide maximum predictions per second, this is the throughput. The throughput may be expressed as the \emph{number of predictions per second}.
    \item \textbf{Data flow applications.} Second, when data samples come sequentially such as in real-time embedded systems, Markov Decision Process (deep reinforcement learning)...  the latency is measured as the \emph{number of milliseconds to get one prediction}.
    \item \textbf{Irregular batch applications.} The third scenario is when the data samples come to an irregular frequency such as web service receiving requests from multiple clients, in this case, the batch computing must be adapted to the client requests characteristics \cite{clipper:2017} and the metric carefully weighted between the latency for reactiveness and the throughput when the service is under a heavy workload of requests.
\end{itemize}

The throughput results on machine A and machine B are presented in appendix~\ref{app:throughput} and figure~\ref{fig:througB}. We decided to not show the latencies to predict 1 single data (data flow applications) because the results were correlated with the throughput with the batch of size 1. No bar in the figure means out-of-memory (VGG with TensorRT) or error compilation (EfficientNet with OpenVINO).

\subsection{Memory consumption}

Lower memory consumption allows for managing more deep learning models and higher complexity ones. The memory consumption is expressed as how many the neural network took on the disk plus the current batch of features flowing through layers. To make global statistics we measure the ensemble and show it in appendix~\ref{app:mem}.

\subsection{Power consumption}

Power consumption is an important metric that interests scientists and companies scientists for diverse reasons such as limiting ecological footprint, energy bills, and reducing thermal problems... We express here the power consumption as watt-seconds of the inference system to predict a fixed amount of $D$ data samples, with $D=4096$. Like the throughput, the value depends on the considered neural network, the chosen inference engine, and the batch size. Then, these measures may be converted in equivalent in CO2 equivalent \footnote{\url{https://www.bilans-ges.ademe.fr}}, energy bill or thermal dissipation...
We express the power consumption $E$ here as watt/sec to predict a fixed amount of data samples $D$, we fix here $D=4096$. $E$ is described with the equation \ref{eq}, with $W$ the mean instantaneous power consumption (watts) during the prediction time $T$ (sec.).

\vspace{-0.3cm}
\begin{equation}
    E=D \times W \times T
    \label{eq}
\end{equation}
\vspace{-0.3cm}

Appendix~\ref{app:power} shows the correlation between the throughput of an inference system and its power consumption. The trend seems to follow the power law with two different coefficients for different GPU generations.
%TODO parler des 2 phenomenes

%The throughput is the number of predictions per second when a large number of data samples are available. In all experiments, we measure the data samples to predict 4096 data samples. A larger number of data samples would increase our benchmark time, but lower data samples would increase the random effect in our measures. Because the batch size may depend on the application of the characteristics of the incoming request, we measure the throughput at different batch size values ranging from 1 to 1024. 1024 may be a constraining factor for the latest GPU devices and further investigations should evaluate greater batch size values.

% latency
%The \textbf{latency} measured as . Because latency may vary due to hardware and operating system reasons we are also interested in the distribution of the latency. We predict sequentially 1000 data samples (i.e, batch size=1) and we measure the 50th, 95th, and 99th percentile of the latency each time.

In terms of power consumption, there is a connection between two phenomena: 
\begin{itemize}
    \item Phenomenon (A). The faster a workload is finished, the less time it has to consume power (kWh).
    \item Phenomenon (B).  The faster a workload is finished, the more hardware piece has been exploited at the same time (cores, memory exchanges). Therefore, more instantaneous energy is consumed by the GPU (watt).
\end{itemize}
This benchmark reveals that phenomenon (A) is of greater importance than B. In this case, faster means more power efficiency.

When an inference framework runs an optimized graph, it has less time to consume power but, on the opposite, it uses more pieces of hardware in parallel and consumes more energy (watt). 
Overall, our experiments on GPUs show that accelerating computing is generally beneficial for power consumption.
%Those 18 experiments spread on two different GPUs show the trend of speed and energy consumption following are lined a whatever is the framework.

\subsection{Loading time}

The loading time is the elapsed time from the neural network representation on the disk and its state in memory where it is ready to predict. Some applications may require a fast service setup to satisfy peak demand periods (e.g., elastic cloud service \cite{elasticity}). In all our benchmarks, the cache system is enabled or implemented to measure the loading time.

The loading times of the ensemble on machine A in ascending order: OpenVINO (3.4 sec.), TensorRT(3.6), ONNX-RT GPU(5.7), ONNX-RT CPU(8.7), Tensorflow CPU (8.7), TensorFlow XLA GPU (42.5). There are loading times for machine B: ONNX-RT CPU (0.5), ONNX-RT GPU (2.1), Tensorflow CPU (2.9), TensorRT (3.9), TensorFlow XLA GPU (29.2).

The load time is hardware-dependent and at first glance difficult to estimate. If it is the key performance metric, performing multiple experiments to select the best framework is unavoidable.
Tensorflow with XLA is especially slow due to the fact there is currently no way to cache the optimized graph on the disk. Removing XLA optimization divides by 6 the loading time, but hurts significantly the prediction time.

\section{Conclusion}
\label{sec:conclusion}

Inference framework is an active field due to the increasing number of applications and the current room for speed improvement.

% Frameworks GPU
%Regarding GPU technologies there is no clear winner between TensorRT, Tensorflow, and ONNX-RT, and selecting the best one will depend on the application scenario and the underlying hardware.

%\emph{ONNX-RT is systematically the fastest in terms of latency and low batch size throughput}. TensorRT is faster on large batch size values such as 512 and 1024 but it may perform poorly on wide convolutions. TensorRT is also the framework that consumes the least memory. %It appears performant  The current state of the art in computer vision seems to follow the path of a smaller but larger number of layers more efficient and sparser neural networks with multiple interconnections between nodes.

% Frameworks CPU
%If we look at CPU optimization, Intel OpenVINO on machine A is always faster than the others compared to ONNX-RT CPU, and Tensorflow-CPU and MLIR showed promising results even if it is still under development. Intel OpenVINO is also well-optimized to save memory compared to Tensorflow CPU.

%Finally, these experiments show the need for different frameworks for training and inference because they require different representations and optimizations. 
As expected, the more optimized and specialized a framework for inference and the more efficient it is. However, ONNX-RT is faster for small batches and fits data flow applications and/or rare requests. TensorRT is faster for throughput and large batch size, it is a good fit for servers under a heavy workload of requests. Tensorflow with XLA  (Accelerated Linear Algebra) allows for simplifying the software stack by re-using the same framework for both training and inference.   %Therefore, the best framework depends on the targeted application.
%Finally we observe that high-level optimization affects more GPU computing than CPU computing, for example, TensorFlow XLA has a low or no impact on the CPU  but improves by about 20\% the throughput in GPU (it also multiplies by 6 the loading time). The different operations fusing that OpenVINO and ONNX-RT CPU exposes may improve the speed to only a small amount (from 0\% to +5\% faster), while ONNX-RT GPU significant gains speed (+12\%).

%TODO parler des ensembles

% Future research
Other post-training optimization consists in reducing the complexity of the graph using calibration data to control and preserve the prediction quality. It includes pruning \cite{pruning} \cite{compressentropy:2020}, distilling \cite{distilling:2018} and lower precision arithmetic \cite{quantiz2011} \cite{binary:2015}. We leave them for future benchmarks and researches.

% TODO conclusion MLBench, MLPErfin, ...

Codes are available on GitHub: \url{https://github.com/PierrickPochelu/inference_framework_benchmark}

\bibliography{bib_pierrick,bib_automl,bib_ensemble,bib_hpml,bib_infer,bib_rl}
\bibliographystyle{IEEEtran}

\appendices
\counterwithin{figure}{section}
\onecolumn

\clearpage
\section{Throughput}
\label{app:throughput}

\begin{figure}[h]
        %\centering
        \includegraphics[width=0.9\linewidth]{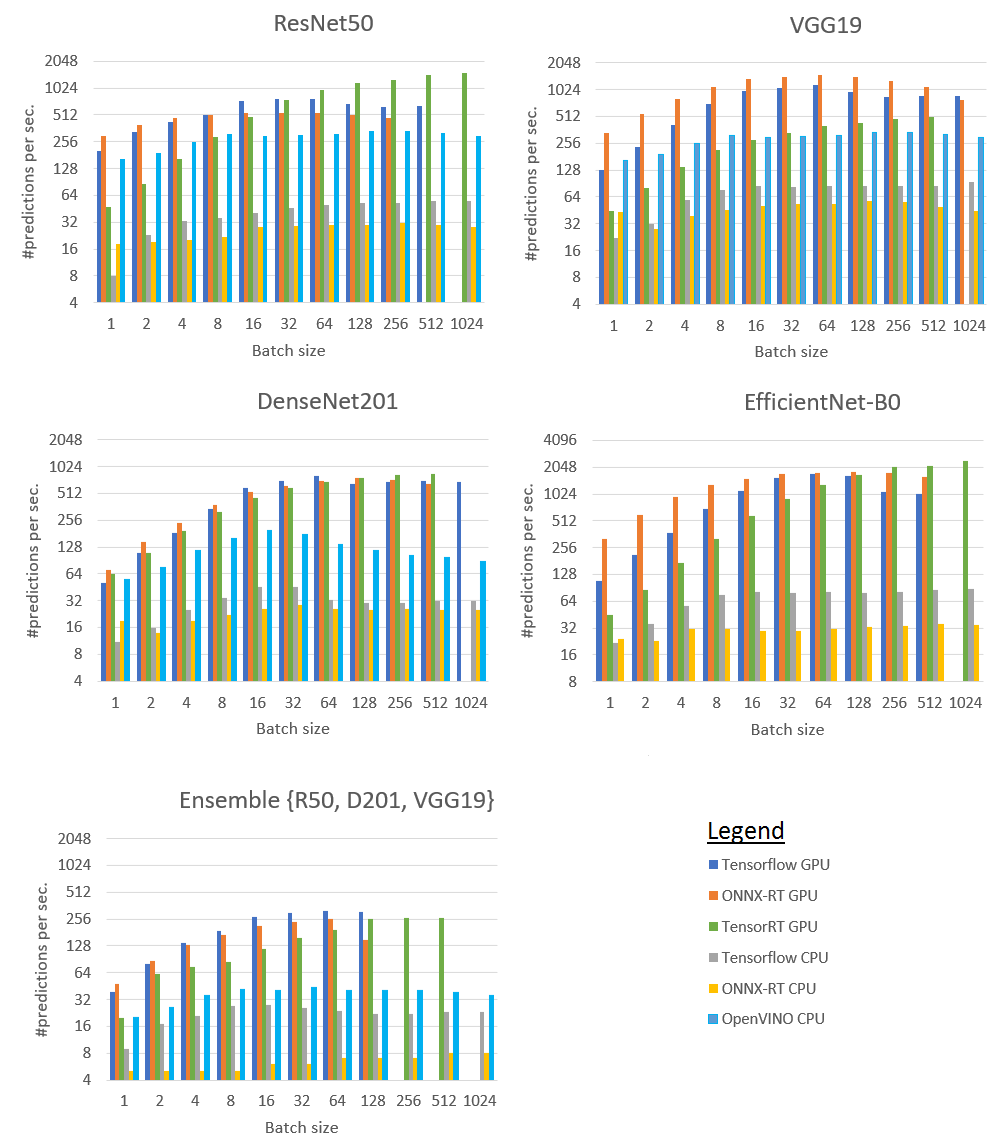}
        \caption{Machine A.}
        \label{fig:througA}
\end{figure}

\begin{figure}[h]
        \includegraphics[width=0.9\linewidth]{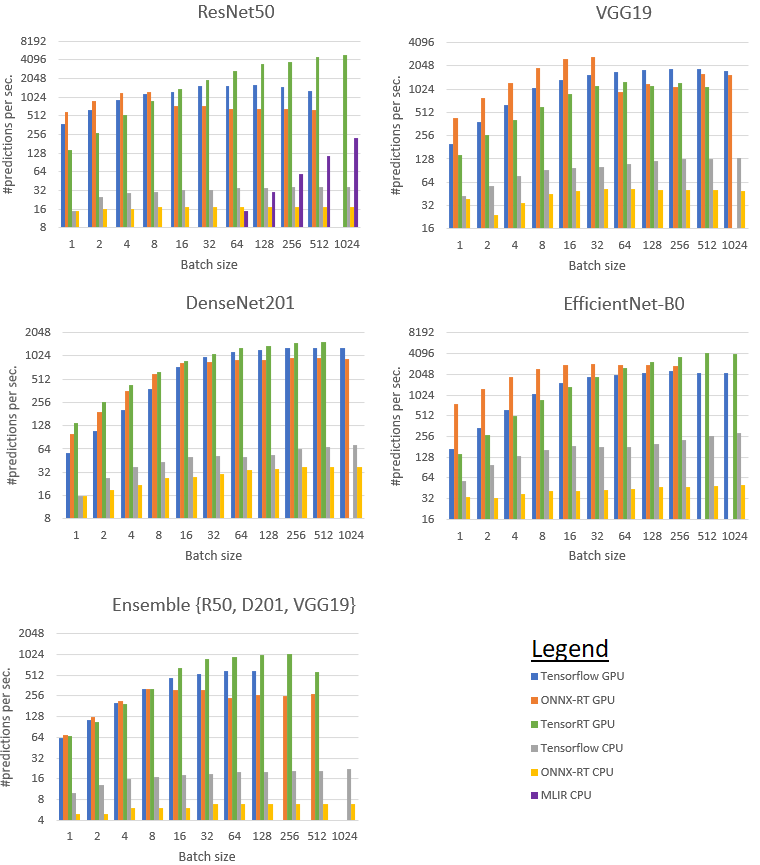}
        \caption{Machine B.}
        \label{fig:througB}
\end{figure}

Figure~\ref{fig:througA} \ref{fig:througB} show the performance of the inference frameworks by varying the batch size and the model. Notice the vertical axis is log2-scale.

%\TODO NOT THE SAME SIZE. THIS IS UGLY.

\clearpage
\section{Memory consumption}
\label{app:mem}
\begin{figure}[h]
        \centering
        \includegraphics[width=\linewidth]{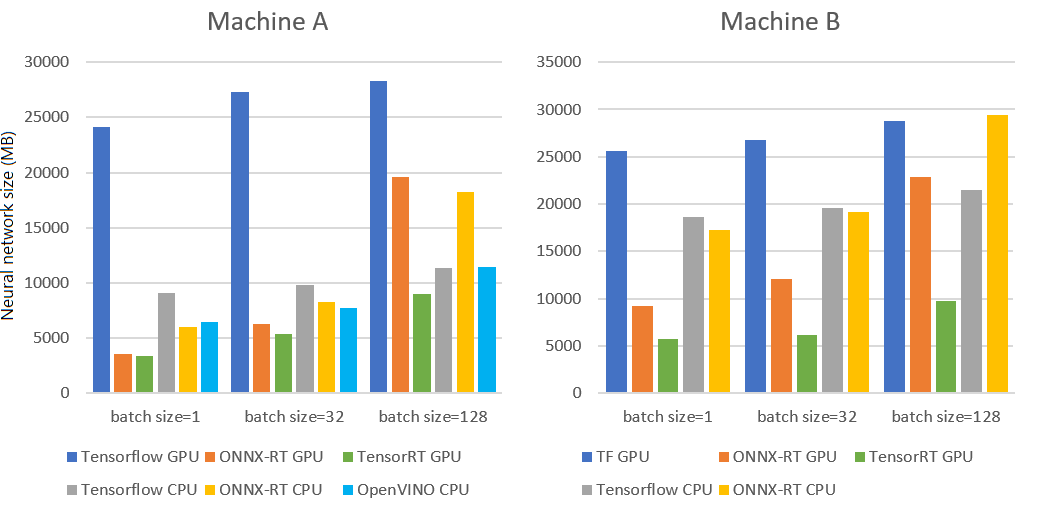}
        \caption{Memory consumption (MB) of the ensemble with different frameworks varying the batch size. }
        \label{fig:memor}
\end{figure}

Figure~\ref{fig:memor} shows the memory consumption of the ensemble \{VGG19, ResNet50, DenseNet201\}. We use this ensemble due to the diversity of topology between neural networks: VGG has wider convolutions, Resnet is deeper, and Densenet contains many jumps between layers.

\clearpage
\section{Power consumption}
\label{app:power}

\begin{figure}[h]
        \centering
        \includegraphics[width=\linewidth]{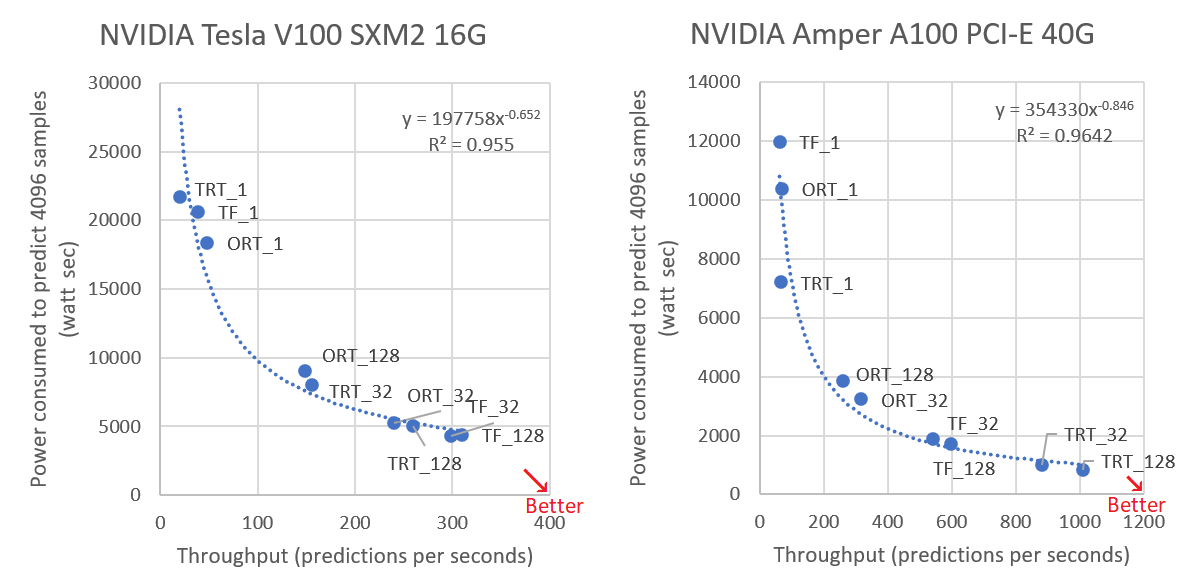}
        %\caption{Plot of the throughput and power consumption of 3 GPU inference frameworks (``TRT'' for TensorRT, ``TF'' for Tensorflow, ``ORT'' for ONNX-RT) with \{1,32,128\} batch size. }
         \caption{Plot of the throughput and power consumption }
        \label{fig:pow}
\end{figure}

Figure~\ref{fig:pow} presents the power consumption (Watt/sec.) of different inference frameworks  (``TRT'' for TensorRT, ``TF'' for Tensorflow, ``ORT'' for ONNX-RT) with \{1,32,128\} batch size. The model running is the ensemble \{VGG19, Resnet50, DenseNet201\}.
%\vspace{-0.5cm}

We use the below command to measure power draw:
\begin{verbatim}
nvidia-smi -i ${GPUID} --format=csv --query-gpu=power.draw --loop-ms=3000
\end{verbatim}
with \${GPUID} the corresponding GPU identifier hosting the neural network.

\end{document}